\documentclass[10pt,twocolumn,letterpaper]{article}

\usepackage[pagenumbers]{wacv} 

\usepackage{graphicx}
\usepackage{amsmath}
\usepackage{amssymb}
\usepackage{booktabs}

\usepackage{graphics}
\usepackage{multirow}
\usepackage{wrapfig}
\usepackage{pifont}
\usepackage{color}
\usepackage{colortbl}
\usepackage{caption}

\usepackage{cuted}
\usepackage{capt-of}

\usepackage[pagebackref,breaklinks,colorlinks]{hyperref}

\usepackage[capitalize]{cleveref}
\crefname{section}{Sec.}{Secs.}
\Crefname{section}{Section}{Sections}
\Crefname{table}{Table}{Tables}
\crefname{table}{Tab.}{Tabs.}

\def\wacvPaperID{*****} 
\def\confName{WACV}
\def\confYear{2025}

\begin{document}

\title{Through the Curved Cover: Synthesizing Cover Aberrated Scenes with Refractive Field}

\author{Liuyue Xie\\
Carnegie Mellon University\\
{\tt\small liuyuex@andrew.cmu.edu}
\and
Jiancong Guo\\
Google\\
{\tt\small jiancong@google.com}
\and
L\'aszl\'o A. Jeni\\
Carnegie Mellon University\\
{\tt\small laszlojeni@cmu.edu}
\and
Zhiheng Jia\\
Google\\
{\tt\small lukezjia@google.com}
\and
Mingyang Li\\
Google\\
{\tt\small mingyangli@google.com}
\and
Yunwen Zhou\\
Google\\
{\tt\small verse@google.com}
\and
Chao Guo\\
Google\\
{\tt\small chaoguo@google.com}
}






\maketitle


\begin{strip}\centering
\date{\includegraphics[width=0.7\linewidth, trim={0 0 0 0}]{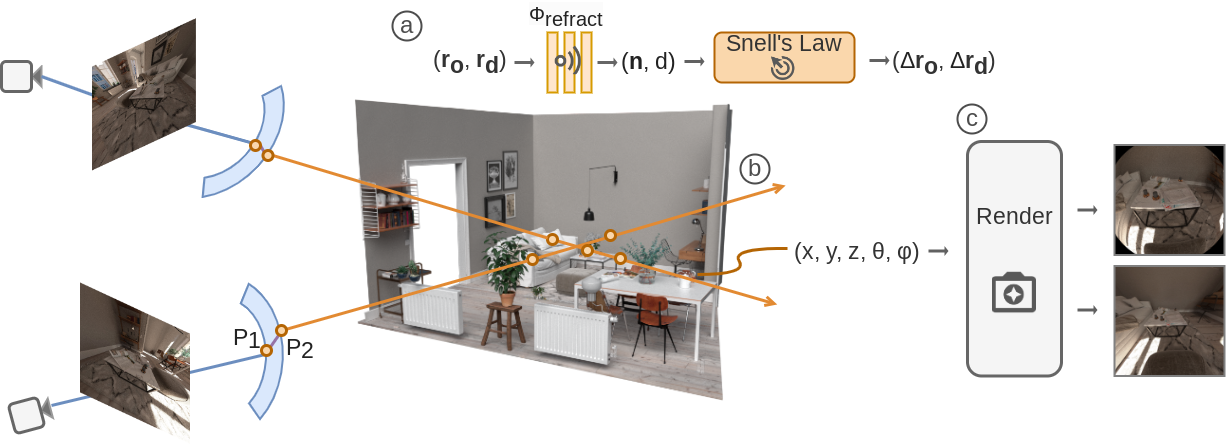}}
\captionof{figure}{This figure illustrates the four modules of the proposed framework: (a) Refractive Field: Estimates surface normals and ray-to-cover distances, using Snell's Law to calculate ray offsets. (b) Ray Sampling: Follows refracted paths, calculating positional offsets for rays.
(c) Radiance Field Rendering: Renders the scene using the sampled points.}
\label{main_pipeline}
\end{strip}

\begin{abstract}
Recent extended reality headsets and field robots have adopted covers to protect the front-facing cameras from environmental hazards and falls. The surface irregularities on the cover can lead to optical aberrations like blurring and non-parametric distortions. Novel view synthesis methods like NeRF and 3D Gaussian Splatting are ill-equipped to synthesize from sequences with optical aberrations. To address this challenge, we introduce SynthCover to enable novel view synthesis through protective covers for downstream extended reality applications. SynthCover employs a Refractive Field that estimates the cover's geometry, enabling precise analytical calculation of refracted rays. 
Experiments on synthetic and real-world scenes demonstrate our method's ability to accurately model scenes viewed through protective covers, achieving a significant improvement in rendering quality compared to prior methods. We also show that the model can adjust well to various cover geometries with synthetic sequences captured with covers of different surface curvatures. To motivate further studies on this problem, we provide the benchmarked dataset containing real and synthetic walkable scenes captured with protective cover optical aberrations. 
\end{abstract}

\section{Introduction}
\label{sec:intro}

The deployment of extended reality (XR) devices in commercial settings depends on their ability to perceive their surroundings accurately and reliably. Recently, XR headsets have been equipped with translucent covers over their front-facing cameras to enhance durability and safety. However, these protective covers present challenges for novel view synthesis due to surface irregularities that can cause image aberrations, leading to degraded synthesis quality. We propose SynthCover, a novel framework for neural view synthesis through refractive surfaces, which addresses this challenge by providing a geometric estimation of surface irregularities that is jointly optimized with the synthesis process, enabling high-quality rendering even from cover-aberrated captures.

Image aberration has been a persistent issue in photography and imaging, particularly for 3D reconstruction. Common optical aberrations, such as blurring and distortions, occur even with monochromatic light sources. To mitigate these effects, specialized blind deblurring algorithms are often used to preprocess and restore image quality, while distortion correction typically relies on physical camera model approximations \cite{Ferreira2005StereoScene, Schonberger2016Structure-from-MotionRevisited, Alterman2017TriangulationDistortions, Faugeras1992CameraExperiments, Hartley1997KruppasMatrix, Pollefeys1999StratifiedConstraint}, using parametric models to realign distorted pixels. Photogrammetry and novel view synthesis frameworks usually derive these parameters using Structure-from-Motion (SfM). However, these methods often struggle to generalize across the non-symmetrical distortions introduced by curved covers \cite{ Mildenhall2020NeRF:Synthesis
, Ramalingam2017ACalibration, Zeller1996CameraRevisited}, a design feature increasingly common in modern XR headsets and dome cameras used in field robotics.

Surface irregularities from the polishing process of these covers can deviate from the intended geometry, causing irregular distortions and image blurring. While high-precision interferometers can measure these irregularities in a lab environment before deployment, scaling this measurement process for mass production and after deployment remains a major hurdle. Synthesizing novel viewpoints from aberrated image sequences is therefore non-trivial, as the surface irregularities introduce non-radially symmetric distortions and optical artifacts that conventional calibration techniques struggle to handle \cite{Xian2023NeuralModeling, Scaramuzza2006AMotion, Garrido-Jurado2016GenerationProgramming, Debevec1997RecoveringPhotographs, Mildenhall2020NeRF:Synthesis
, Xiong2021In-the-WildSurfaces}. These distorted features cannot be reliably matched to their warped counterparts, especially in complex scenes \cite{Yaldiz2021DeepFormableTag:Markers
, Garrido-Jurado2016GenerationProgramming, Kim2008RobustCorrection, Hartley1997KruppasMatrix, Pollefeys1999StratifiedConstraint}. Recent advancements have explored embedding camera parameter tuning within reconstruction objectives, allowing for synthesis with noisy calibration parameters \cite{Jeong2021Self-CalibratingFields, Xian2023NeuralModeling} and reducing dependency on precise initializations. However, these methods struggle when camera rays deviate significantly from the assumed camera model, which is often the case with curved covers.

To address these limitations, we introduce SynthCover, an innovative framework that explicitly models the cover's geometry and adjusts the camera rays accordingly to account for optical aberrations. SynthCover effectively handles the complex distortions introduced by curved covers, surpassing the limitations of conventional camera model-based methods. By learning the cover geometry, our framework adds an additional degree of freedom for ray propagation, enabling an accurate representation of cover-induced distortions while adhering to physical refractive laws.

Our framework demonstrates superior performance compared to state-of-the-art methods, including camera-calibrating novel view synthesis frameworks, in rendering cover-aberrated sequences from real-world and simulated captures. We demonstrate that our cover geometry modeling approach effectively handles ray distortions in complex scenes, producing results superior to existing novel view synthesis frameworks \cite{Jeong2021Self-CalibratingFields, Xu2023VR-NeRF:Spaces, Kerbl20233DRendering} for modeling camera-aberrated sequences. Our contributions can be summarized as follows:
\begin{itemize}
\itemsep-0.25em
\item End-to-end novel view synthesis for cover-protected cameras.
\item Estimating protective-cover's surface figure geometries through ray tracing.
\item Protective cover aberrated walkable scene captures for indoor and outdoor environments.
\end{itemize}

We note several assumptions in our work for estimating the surface geometries. Surface geometry estimation in optics usually involves considering spherical aberration, coma, astigmatism, curvature of field, and distortion. Since images, once captured, contain compressed light information and thus incomplete ray characteristics, we here simplify the modeling to handle two general categories of aberration: image blurring and distortions. We assume a monochromatic incoming light compatible with the ray tracing representation in the novel view synthesis methods. The material dispersion model is assumed to be constant with wavelength. 

\begin{figure*}[!t]
\centering
\includegraphics[width=0.8\linewidth, trim={0 0 0 0}]{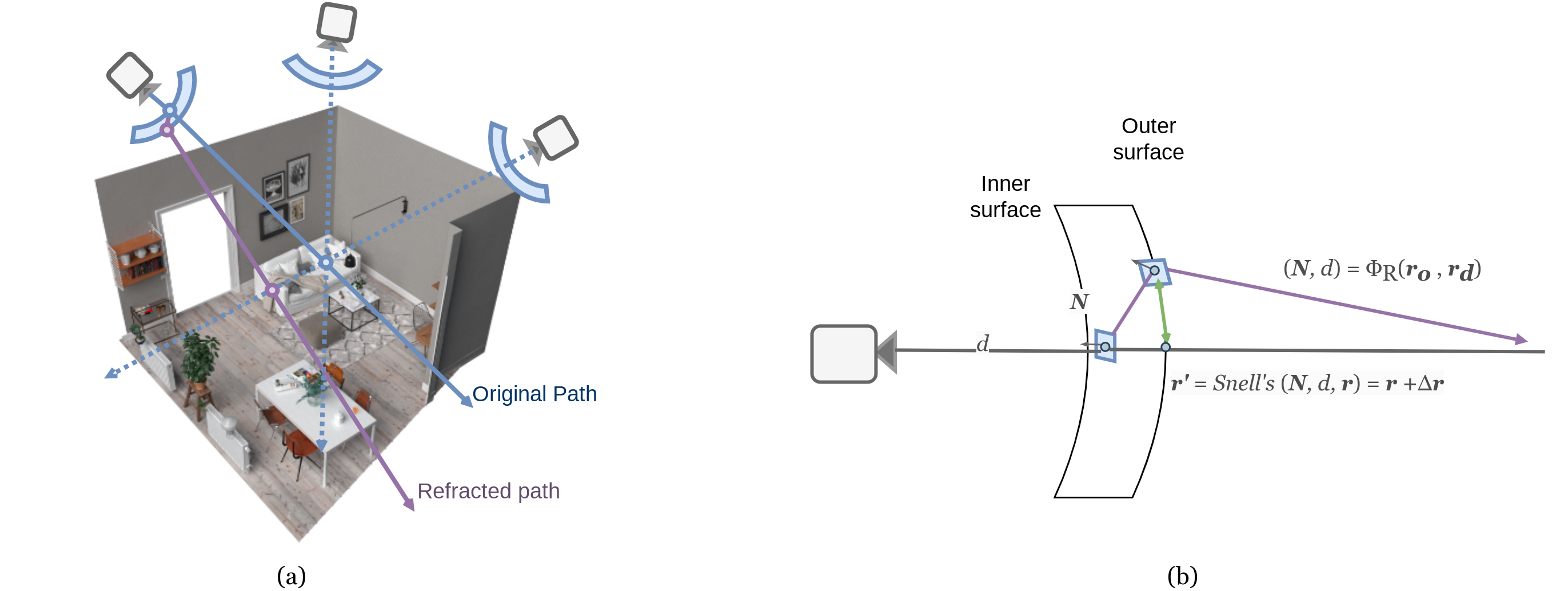}
\caption{(a) Refraction from curved cover distorts light paths and causes rendering artifacts. (b) Our method addresses this by modeling the cover geometry and analytically bending the rays conforming to Snell's physical refraction law.}
\label{pipeline}
\end{figure*}

\section{Related Works}
\label{sec:citations}
\textbf{Camera distortion modeling and optical aberrations.}
The protective covers result in geometric optic aberration in the form of distortion, and visual aberration perceived as image blurring. In the past, camera distortion has been addressed in computer vision through the use of parameterized models tailored to specific camera types, as evidenced by a range of studies \cite{Ramalingam2005TowardsCalibration, Faugeras1992CameraExperiments, Grossberg2001AParameters, Sturm2003AAlgorithms, Zeller1996CameraRevisited, BROWNDC1971Close-Calibration, Kim2008RobustCorrection, Nister2005Non-parametricSelf-calibration, Scaramuzza2006AMotion, Kilpela1981CompensationCoordinates}. These models have shown good generalization across various capturing devices. However, they fall short in accurately representing per-pixel distortions, especially with image blurring, leading to the exploration of more flexible calibration methods that can accommodate these nuances for novel view synthesis, which relies on accurate ray-to-pixel associations \cite{Xian2023NeuralModeling, Fitzgibbon2001SimultaneousDistortion, Jeong2021Self-CalibratingFields}.

Recent advances in camera modeling have focused on estimating per-pixel distortions, addressing the limitations of traditional camera models, and accommodating sensors that deviate from conventional models \cite{Pollefeys1999StratifiedConstraint, Bergmann2018OnlineSLAM, Xiong2021In-the-WildSurfaces, Schops2020WhyTwelve}. Subsequent research has extended these concepts, proposing advanced optimization strategies for imaging models that transform 3D rays into pixels and exploring the impact of different perspective distortion effects on the image plane \cite{Fitzgibbon2001SimultaneousDistortion, Dunne2010EfficientProjection, Sturm2003AAlgorithms, Bousaid2020PerspectiveMeasurements}. While these per-pixel calibration methods offer promising flexibility and generality, they introduce a significant number of parameters, often relying on sensor-specific knowledge \cite{Scaramuzza2006AMotion, Grossberg2001AParameters, Schops2020WhyTwelve, Claus2005ACameras, Kilpela1981CompensationCoordinates, Camposeco2015Non-parametricCameras}, which complicates the optimization process and makes them less suitable for integration into novel view synthesis pipelines. In contrast to these approaches, our method departs from hand-crafted parameterization of camera models, focusing instead on per-ray estimation across generic captures. 

\textbf{Self-calibrating novel-view synthesis.}
Neural Radiance Fields (NeRF) have revolutionized the synthesis of novel views by leveraging a coordinate network to implicitly learn a volumetric scene function \cite{Mildenhall2020NeRF:Synthesis}. Subsequent enhancements to NeRF have been proposed, focusing on various aspects of volumetric rendering and scene reconstruction for more accurate surface and scene geometry representations \cite{Barron2021Mip-NeRF:Fields, Xu2022Point-NeRF:Fields, Xu2023VR-NeRF:Spaces, Attal2022LearningEmbedding}. Gaussian Splatting was subsequently proposed to provide an explicit alternative to modeling the scenes with an improved rendering speed and training efficiency. 

While initially focused on perspective image sequences, recent works have extended NeRF to distorted image captures \cite{Jeong2021Self-CalibratingFields, Xian2023NeuralModeling}. These approaches introduce an optimizable camera model that learns parameter offsets to the initialized parameters, enabling the rendering when SfM techniques like COLMAP \cite{schoenberger2016sfm} fail to accurately initialize camera parameters. However, these models do not specifically account for refractive mediums, which limits their effectiveness in rendering scenes looking through covers. 3DGS, on another hand, does not propagate gradients to camera intrinsic parameters as it relies on perspective rasterization. In the context of extended reality, device captures, we propose to embed cover-aberrated refraction modeling to achieve end-to-end learning on real-world captures, enabling both NeRF and 3DGS to account for the complex distortions introduced by the curved cover, leading to more flexible synthesis for challenging mobile captures.
\section{Preliminary on Novel View Synthesis}
Neural Radiance Field (NeRF) \cite{Mildenhall2020NeRF:Synthesis} and 3D Gaussian Splatting (3D-GS) \cite{Kerbl20233DRendering} are distinct 3D scene synthesis techniques from 2D images. NeRF's implicit volumetric rendering Gaussian Splatting's explicit $\alpha$-blending operate on a similar principle. For NeRF, with color $C$, density $\sigma$, transmittance $T$, and point sampling interval $\delta$, the color is produced with volumetric rendering along a ray as:
\setlength{\belowdisplayskip}{4pt} \setlength{\belowdisplayshortskip}{4pt}
\setlength{\abovedisplayskip}{1.5pt} \setlength{\abovedisplayshortskip}{1.5pt}
\begin{equation}
    C = \sum^N_{i=1} T_i (1 - exp(-\sigma_i\delta_i))\boldsymbol{c_i}, \text{ with }
    T_i = exp(-\Sigma^{i-1}_{j=1}\sigma_j\delta_j).
\end{equation}
As an explicit approach, Gaussian Splatting instead computes the color blending from $N$ ordered points overlapping each pixel with their respective $\alpha$ values as:
\begin{equation}
    C = \Sigma_{i\in N} \boldsymbol{c_i} \alpha_i \prod^{i-1}_{j=1} (1-\alpha_j).
\end{equation}

For both approaches, the obtained color is contrasted against the corresponding color in the target image to refine the model's parameters. Both techniques are adept at synthesizing three-dimensional scenes, offering comparable degrees of signal supervision to fine-tune ray deformations \cite{Mildenhall2020NeRF:Synthesis, Kerbl20233DRendering}. Pose matrices $M = [R, T]$ and projection matrix $P$ governing the outward ray directions are required to project a learned 3D canonical volume to the 2D image plane for rendering the colors. SfM methods have previously been used to provide camera estimates, but they fall short with aberrated captures. We address this by learning the protective cover geometries and accordingly correcting the outward rays. This ensures that the outward camera rays adhere to the aberrated renders, mitigating the issues of inaccurate canonical space to camera plane projections. 

\begin{figure*}[!htbp]
\centering
\includegraphics[width=\linewidth, trim={0 0 0 0}]{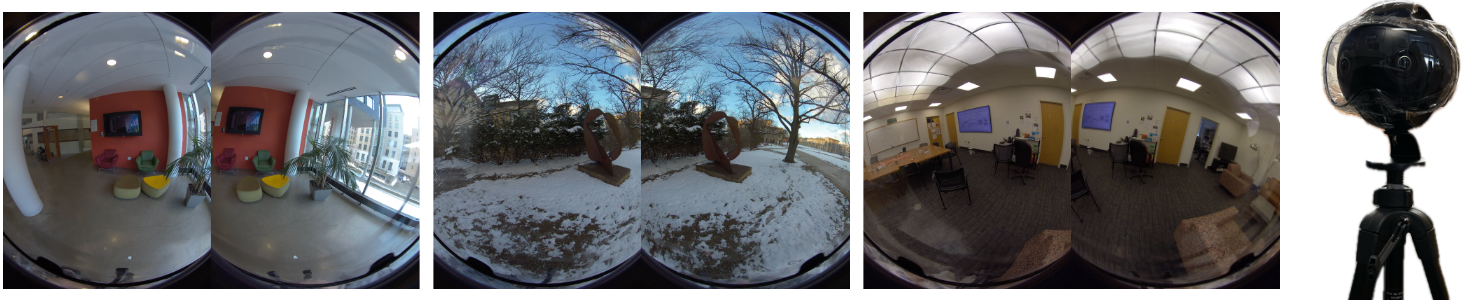}
\caption{Sample images from the CurvedCover dataset and the capturing rig.}
\label{capture}
\end{figure*}

\section{Modeling Optical Aberrations from Cover Geometry}
We study the problem of novel view synthesis from a monocular cover-aberrated video sequence. The surface figure of the translucent protective cover induces image blurring and distortions, we model these optical irregularities explicitly with a Refractive Field such that the rays coming out from the camera assembly reflect the optical surface geometry. Using the Structure-from-Motion method COLMAP, we obtain an initial estimation of camera poses, intrinsic parameters, and distortion coefficients. The camera intrinsics establish the basic geometric properties of the camera, determining the overall projection behavior of the rays, while the distortion parameters introduce non-linear modifications to the rays, bending them in a way that deviates from ideal pinhole projection. Lifting the plane coordinates $\boldsymbol{x}$ to the corresponding ray $\textbf{r}$, the rays originating from each camera's viewpoint can be expressed as:
\begin{equation}
    \begin{split}
        \boldsymbol{r} = f^{-1}(\boldsymbol{x}) = \boldsymbol{r}(d) = \boldsymbol{r}_o + d \cdot \boldsymbol{r}_d, \\
    \end{split}
\end{equation}
where $f(\boldsymbol{r})$ derived from the camera intrinsic and distortion parameters denotes the mapping of a 3D ray $\boldsymbol{r}$ to a 2D image plane coordinate $\boldsymbol{x}$. $\boldsymbol{r_o} \in \mathbb{R}^{n}$ and $\boldsymbol{r_d} \in \mathbb{R}^{n}$ respectively denote the ray's origin and direction, and $d$ is the ray's travel distance. Assuming that the ray's origin is provided by SfM initialization as the camera's world coordinate at a given frame, we seek the optimal outward ray directions $\textbf{r}_d$ such that the rendered image $\hat{\mathcal{I}}(\boldsymbol{r})$ matches the input image ${\mathcal{I}}(\boldsymbol{r})$.

Recognizing the potential inaccuracy of SfM-derived camera intrinsic and distortion parameters, we employ a differential fine-tuning strategy. This process optimizes the parametric camera model until the remaining optical aberrations are solely attributed to the curved cover. 

The curved cover optical aberrations are usually caused by a thick cover's refractions and surface shape irregularities. When the cover surfaces depart from flatness, a collimated beam would expand or focus and lead to deviations from rectilinear projections. While intuitively, the non-parametric per-ray aberrations can be modeled by learning ray offsets with a ray Refractive Field, directly predicting the distortions with each individual ray is inherently ill-posed since it essentially demands the network to implicitly understand both the geometry of the medium and the physical behavior of Snell's refractive law. In our proposed SynthCover model, we instead decouple the learning objective of the Refractive Field by modeling the physical geometry of the refractive medium with a network and then analytically bending the lights according to the learned geometry through Snell's Law. 

Decoupling the Refractive Field objective alleviates the network's learning demand and adds transparency to the optimization. The mathematical formulations of the differentiable parametric camera model, Refractive Field network, ray sampling, and regularizations are detailed in the following sections. The implementation details are provided in Appendix Sec. \ref{implementation}.
\subsection{Differentiable Parametric Camera Model}
We adopt a differentiable Brown-Conrady model to describe the camera without protective cover aberrations. We note that when the optical power induced by the protective cover is zero, the capturing system would resort to Brown's camera model. The camera model consists of intrinsic parameters describing the optical center and focal lengths, as well as distortion parameters describing the deviations from an ideal pinhole camera. Directly learning the camera intrinsics $\boldsymbol{K}$ leads to a highly non-convex optimization problem \cite{Jeong2021Self-CalibratingFields}. We instead decompose the camera intrinsics into an initialized $\boldsymbol{K}_o$ matrix and learnable offset parameter matrix $\Delta\boldsymbol{K}$ such that the initialized intrinsics reduce the local minima that can compromise the learning process. The revised intrinsic matrix can thus be expressed as $\boldsymbol{K}  \in \mathbb{R}^{3 \times 3} = \boldsymbol{K}_o + \Delta \boldsymbol{K}$ where the norm of $\Delta \boldsymbol{K}$ would be bounded. Similarly, the distortion model $\mathcal{D}$ consists of the initial parameter estimates and offsets to be refined through optimization. 

The camera model describes the mapping between points on the camera sensor $\boldsymbol{x} \in \mathbb{R}^{2 \times 2}$ to their 3D locations $\boldsymbol{X} \in \mathbb{R}^{3 \times 3}$. To express a camera with a ray bundle, we define the ray origins as the camera's location in the world, and we lift the points from the camera sensor to their locations in the world coordinate, then subtract the camera centers to get the ray directions. Deriving the lifting from pixels to rays, the rays in the camera coordinate system are expressed as $\boldsymbol{r_d}(\boldsymbol{p}) = \mathcal{D}(\boldsymbol{K}^{-1}\boldsymbol{p})$ with $\boldsymbol{r_o} = 0$ as rays originating from the camera's position. The extrinsic $\boldsymbol{R}$ and $\boldsymbol{T}$ can be used to transform the rays from the local camera coordinate to the world coordinate as:
$
    \boldsymbol{r_d} = \boldsymbol{R} \cdot \mathcal{D}(\boldsymbol{K}^{-1}\boldsymbol{p}) + \boldsymbol{T}; 
    \boldsymbol{r_o} = \boldsymbol{T}.
  \label{eq:c2w}
$
Detailed formulation of the differentiable camera model can be found in Appendix Sec. \ref{cam_model}.
\begin{figure*}[ht]
\centering
\includegraphics[width=\linewidth, trim={0 0 0 0}]{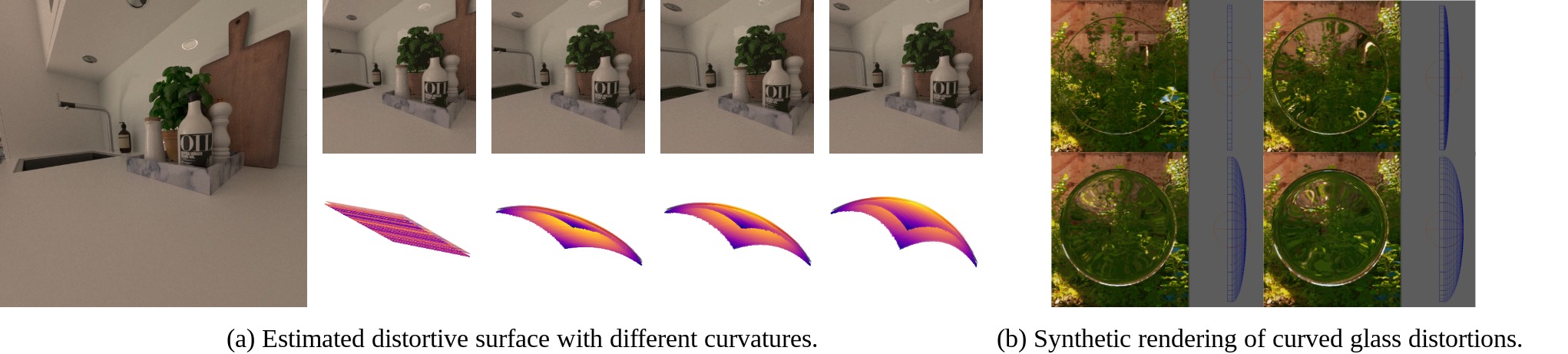}
\caption{(a) Surface reconstruction of the curved cover. (b) Simulation setup for recreating the distortion induced by the curved cover.
}
\label{surface}
\end{figure*}

\begin{table*}[hbt!]
\small
    \centering
    \caption{\textbf{Quantitative comparison on the FisheyeNeRF dataset and EyefulTower dataset.} See Sec \ref{evaluation_section} for an analysis of the performance.}
    \resizebox{\textwidth}{!}{\begin{tabular}{c|ccc|ccc|ccc|ccc}
    \toprule
        method & \multicolumn{3}{c}{globe} & \multicolumn{3}{c}{cube} & \multicolumn{3}{c}{table} & \multicolumn{3}{c}{sofa area} \\
        
        & PSNR $\uparrow$& SSIM $\uparrow$& LPIPS $\downarrow$ & PSNR $\uparrow$& SSIM $\uparrow$& LPIPS $\downarrow$ & PSNR $\uparrow$& SSIM $\uparrow$& LPIPS $\downarrow$ & PSNR $\uparrow$& SSIM $\uparrow$& LPIPS $\downarrow$  \\  \midrule \midrule

        NeRF & 19.3 & 0.472 & 0.731 & 18.5 & 0.534 & 0.672 & 21.5 & 0.788 & 0.333 & 21.7 & 0.672 & 0.364\\
        SCNeRF & 23.8 & 0.598 & 0.633 & 23.2 & 0.605 & 0.616 & 22.9 & 0.728 & 0.275 & 25.1 & 0.769 & 0.348\\
        NeuroLens & \underline{24.5} & \underline{0.724} & \underline{0.148} & \underline{25.3} & \underline{0.716} & \underline{0.163} & \underline{23.4} & \underline{0.817} & \underline{0.260} & \underline{27.8} & \underline{0.837} & \underline{0.189}\\
        Ours & \textbf{28.4} & \textbf{0.730} & \textbf{0.124} & \textbf{27.3} & \textbf{0.728} & \textbf{0.147} & \textbf{27.2} & \textbf{0.833} & \textbf{0.189} & \textbf{29.2} & \textbf{0.859} & \textbf{0.136}\\
    \bottomrule
    \end{tabular}
    }
    \label{table_fisheye_eyeful}
\end{table*}

\begin{table*}[!htbp]
\small
    \centering
    \caption{\textbf{Quantitative comparison on the CurvedCover dataset.} See Sec \ref{evaluation_section} for an analysis of the performance.}
    \resizebox{\textwidth}{!}{\begin{tabular}{c|ccc|ccc|ccc|ccc}
    \toprule
        method & \multicolumn{3}{c}{robot arm} & \multicolumn{3}{c}{scotty} & \multicolumn{3}{c}{statue} & \multicolumn{3}{c}{coffee table} \\
        
        & PSNR $\uparrow$& SSIM $\uparrow$& LPIPS $\downarrow$ & PSNR $\uparrow$& SSIM $\uparrow$& LPIPS $\downarrow$ & PSNR $\uparrow$& SSIM $\uparrow$& LPIPS $\downarrow$ & PSNR $\uparrow$& SSIM $\uparrow$& LPIPS $\downarrow$  \\  \midrule \midrule

        NeRF & 16.1 & 0.711 & 0.8176 & 16.9 & \underline{0.7346} & 0.8682 &  22.1 & 0.6082 & 0.6177& 23.4 & 0.7678 & 0.5977\\
        SCNeRF & 25.9& \underline{0.8761}& \underline{0.2664}& 25.4 & 0.6065& 0.5890 & \underline{26.8} & 0.7187 & 0.3197 &29.5 & 0.8473 & 0.4103\\
        NeuroLens & \underline{26.5} & 0.7799 & 0.3138 & \underline{25.6} & 0.6463 & \underline{0.4455} &26.6 & \underline{0.7897} & \underline{0.2483} & \underline{32.6} & \underline{0.8926} & \underline{0.3280}\\
        Ours & \textbf{28.8}& \textbf{0.9287} & \textbf{0.1564} &\textbf{26.8}& \textbf{0.8972} & \textbf{0.2649} &\textbf{27.3} & \textbf{0.8137} & \textbf{0.2121} &\textbf{32.9} & \textbf{0.9048} & \textbf{0.2873}\\
    \bottomrule
    \end{tabular}
    }
    \label{table_curved_cover}
\end{table*}

\subsection{Refractive Medium Estimation}
We design a network to model the refractive medium $\Phi_R(\boldsymbol{r_o}, \boldsymbol{r_d}) = (d, \boldsymbol{n})$ to estimate the cover surface geometries. The Refractive Field network takes ray direction vectors, $\boldsymbol{r_o}$ and $\boldsymbol{r_d}$, as inputs and then predicts the distance to the two incident surfaces as well as the incident normal. The predicted distances describe the travel distance of each ray to both the cover surfaces. This is estimated to track the incident locations of the rays when the placement of the protective cover and thickness of the cover are both unknown. The outputs from the distortion network are used to analytically compute ray offsets, which are subsequently added to the corresponding rays before they are projected to the world space for rendering. These offsets implicitly account for the distortions introduced by the refractive medium. We leverage an invertible neural network (INN) architecture for this network since it can effectively compute both the forward transformation (distortion) and its inverse (undistortion). This invertibility allows the network to model the complex, two-way relationship between the undistorted and distorted rays.

Since a lens generally has two refractive surfaces, we assume that the rays refract exactly twice, with the first refraction passing the rays from the air to the covered inner surface, and the second refraction passing the rays through the outer surface back to the air. We neglect considerations of self-occlusions on the refractive path to constrain the problem. The estimation of outward rays is detailed as follows. 

For each time step in the sequence, the previously derived rays from the camera are sent to the Refractive Field network $\Phi_R$. The network outputs the travel distance $d \in \mathbb{R}^{n \times 2}$ of each ray to two surfaces $\boldsymbol{S}_1$ and $\boldsymbol{S}_2$ and the corresponding incident normal vectors $\boldsymbol{n} \in \mathbb{R}^{n \times 2}$. With the estimated geometry, a ray $\boldsymbol{r}$ refracts through the first surface at $\boldsymbol{X}_{s1}$ with the estimated incident normal using Snell's Law which governs the angle of light bending when passing across two mediums with different indexes of refraction $(z_1, z_2)$.

We use a vector form of Snell's Law for the refraction computation. In the vector form notation, $\boldsymbol{r} \in \mathbb{R}^3$ denotes the incident ray, $\boldsymbol{n} \in \mathbb{R}^3$ denotes the normal vector, and $\boldsymbol{r}' \in \mathbb{R}^3$ is the transmitted ray. The Snell's Law can then be expressed as:
\begin{equation}
    \begin{split}
    \boldsymbol{r}' = \eta (\boldsymbol{r} + c_1 \boldsymbol{n} ) - c_2 \boldsymbol{n} ,
    \label{eq:snells_vector}
  \end{split}
\end{equation}
where $c_1 = <\boldsymbol{n}, \boldsymbol{r}>$, $c_2 = \sqrt{1-\eta^2 (1 - c_1^2)}$ are two compressed terms, and $\eta = \frac{z_1}{z_2}$ denotes the ratio between two medium's refractive indices. The refracted ray directions through the first stage can thus be computed accordingly, with the rays originating from the cameras defining the incident rays. Similarly, the ray passing through the second surface $\boldsymbol{S}_2$ of the curved cover to exit the refractive medium can be computed again using Snell's Law. The resultant rays outward from the second surface are then used for subsequent rendering. 

Directly estimating the distances and normal vectors from an unconstrained network is ill-posed, since the predicted cover geometry through distance estimation may deviate from the surface geometry described by the normal predictions. To ensure that the network embeds the correct geometry of the cover, we regularize the distance and normal estimations with a consistency loss detailed in Section \ref{losses}.



\subsection{Sampling and Rendering Through Refractive Cover}
The captured scene is distributed along the direction outward from the cover's outer surface. With the surface approximated with the estimated intersection $\boldsymbol{X}_{s2}$, the ray $\boldsymbol{r}'(d)$ beyond the outer refractive surface can be expressed as:
\begin{equation}
    \boldsymbol{r}'(t) = \boldsymbol{X}_{s2} + d \cdot \boldsymbol{r}_d'.
\end{equation}
For each ray in NeRF synthesis, we sample points $\boldsymbol{x}$ along $\boldsymbol{r}'(d)$ then feed the sampled 3D points to the Radiance Field for further prediction of the volumetric scene. With the rays accounting for the refracted offsets, the modified NeRF can now be optimized directly through the aberrated captures. 

Synthesizing 3D Gaussian Splatting (3DGS) scenes requires slight modifications to the optimization as a rasterization method. 3DGS is initialized with a sparse point cloud from SfM, and it necessitates rectifying the training views to align with the perspective projection in the rasterization process. To adapt our approach to 3DGS's formulation, we rectify the training views, modify the training objective to correct residual optical aberrations, and include parameter estimates in addition to the ray offsets. The Gaussians are represented by parameters $(\boldsymbol{\mu}, \boldsymbol{R}, \boldsymbol{S}, \boldsymbol{sh})$, which denote the Gaussian's mean, rotation, scale, and ray-dependent spherical harmonics, respectively. Since the Gaussian parameters, excluding the Spherical Harmonics terms, are not directly associated with the camera ray directions, we introduce two additional layers to capture Gaussian deformations caused by distorted rays. The parameter offsets $(\Delta \boldsymbol{\mu}, \Delta \boldsymbol{R}, \Delta \boldsymbol{S})$ are estimated by the deformation network, while $\Delta \boldsymbol{sh}$ is derived from the ray offsets. These offsets are applied to the canonical Gaussians before rendering, and the rendered results are compared with the rectified image.


\subsection{Regularization}
\label{losses}
We optimize the differentiable camera parameters and distortion field with photometric loss and normal consistency loss. The photometric loss compares the 2D image observations and the rendered results. It builds on the intuition that if the camera forms the images correctly, the color of the rendered pixels from each ray $\boldsymbol{r}$ modeled by the learnable terms should match its projected 3D counterpart in the reference image as:
\begin{equation}
    \mathcal{L}_{photometric} = \Sigma_{\boldsymbol{r}\in \boldsymbol{N}} ||\hat{\mathcal{I}}(\boldsymbol{r}) - \mathcal{I}(\boldsymbol{r})||^2_2
\end{equation}

Besides the photometric differences between the rendered image and the captured image, we also minimize the difference between the predicted normals $\boldsymbol{n_{i}}$ from the Refractive Field and the fitted local normals from the predicted distances $\boldsymbol{n_{d_i}}$. The normal consistency loss is defined as follows:
\begin{equation}
    \begin{split}
    \mathcal{L}_{normals} = \Sigma_{\boldsymbol{r} \in \boldsymbol{N}} ||\boldsymbol{n_{d_r}} - \boldsymbol{n_{r}}||^2_2,
    \label{eq:ray_consistency}
  \end{split}
\end{equation}
where $\boldsymbol{N}$ denotes the collection of ray paths that cross the refractive medium. We fit the local normals in $(3 \times 3)$ neighborhoods from the predicted distances to surfaces $d$ as $\boldsymbol{n_{d}}$ then regularize the network's direct normal estimations  $\boldsymbol{n}$ with the fitted normals. 
\section{Experiments}
\label{sec:result}
In this section, we validate our proposed method on four static scene datasets captured with different levels of camera distortions. We demonstrate the quantitative and qualitative results achieved on the LLFF dataset \cite{mildenhall2019llff}, EyefulTower dataset \cite{Xu2023VR-NeRF:Spaces}, FisheyeNeRF dataset \cite{Jeong2021Self-CalibratingFields}, and our own CurvedCover dataset. Notably, our method achieves state-of-the-art performance across both cameras without covers and those equipped with curved covers. 


\subsection{Evaluation}
\label{evaluation_section}

\begin{figure*}[!hbt]
\centering
\includegraphics[width=0.95\linewidth, trim={0 0 0 0}]{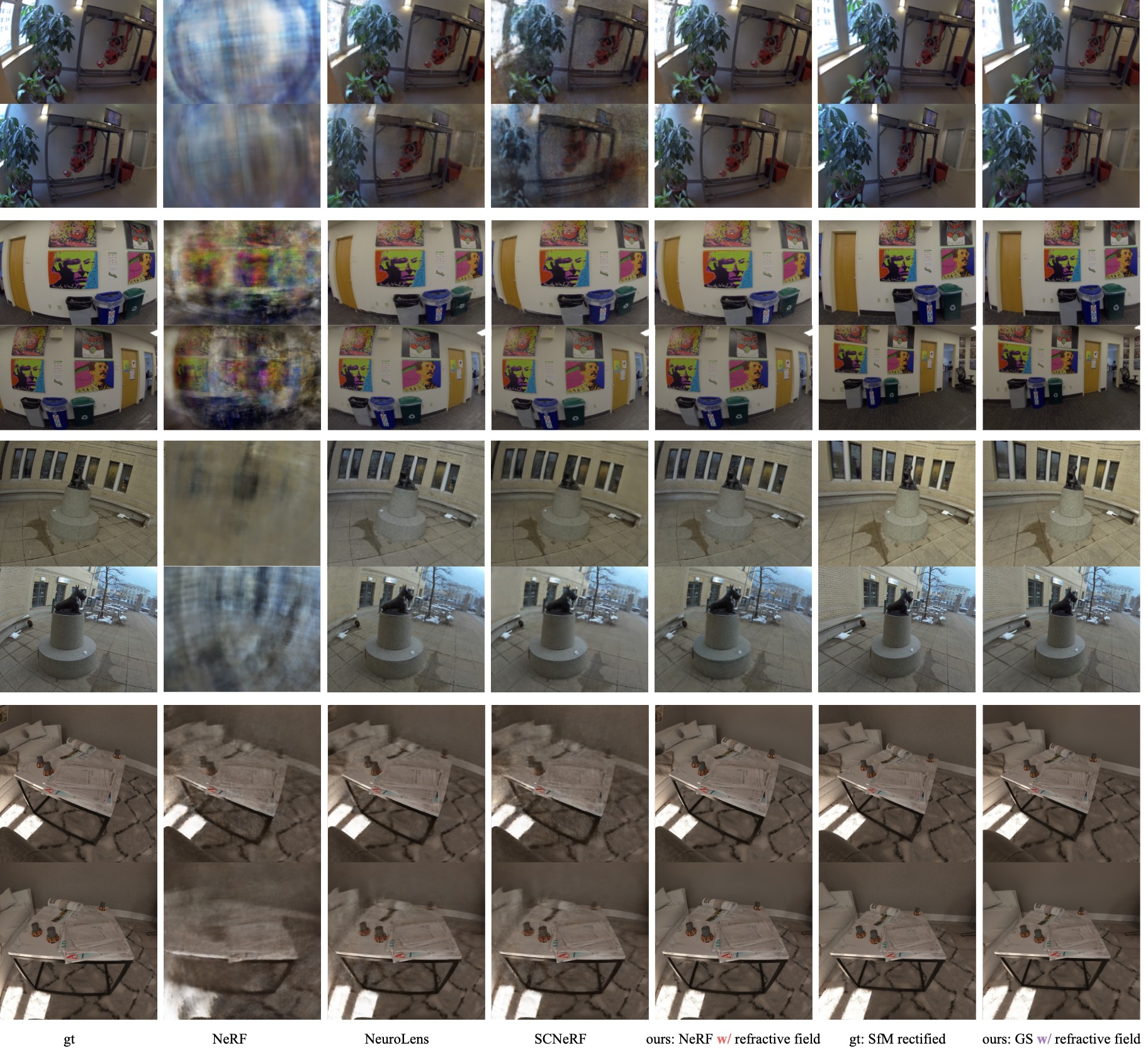}
\caption{This figure compares NeRF variants, including ours. Ground truth images are shown in the first column. All methods use RealityCepture for initial calibration. Our approach shows consistent accuracy across datasets.
}
\label{main_qualitative}
\end{figure*}

This section presents a quantitative and qualitative analysis of our method's performance. Our aim is to investigate whether explicit cover modeling helps disambiguate ray bending paths and enhance the reconstruction of distorted scenes. We also explore whether the proposed camera model potentially hinders the reconstruction of scenes captured without a cover. For implementation details, please see Appendix Sec. \ref{implementation}. 
\begin{figure*}[!t]
\centering
\includegraphics[width=0.95\linewidth, trim={0 0 0 0}]{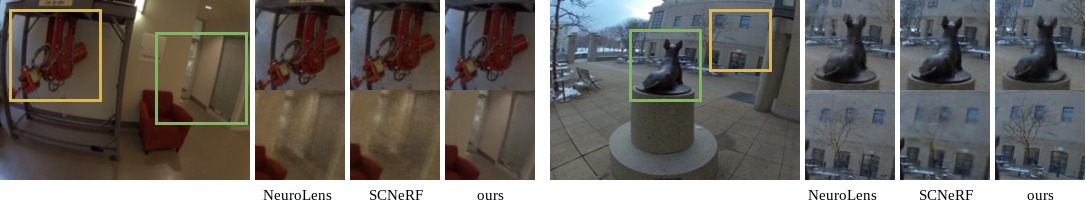}
\caption{This figure compares NeRF variants, including ours. Ground truth images are shown in the first column. All methods use RealityCepture for initial calibration. Our approach shows consistent accuracy across datasets.
}
\label{blowup}
\vspace{-4mm}
\end{figure*}

\textbf{Benchmarked datasets.} We evaluate our method using both real-world and synthetic monocular datasets with varying camera distortions. The real-world datasets include LLFF \cite{mildenhall2019llff} with minimal distortion, EyefulTower \cite{Xu2023VR-NeRF:Spaces} captured with fish-eye camera in office scenes, FisheyeNeRF \cite{Jeong2021Self-CalibratingFields} with wide-angle outdoor scenes, and our CurvedCover dataset captured by dual fish-eye cameras with a curved Apple Vision Pro front-facing cover as shown in Fig. \ref{capture}. We also include synthetic scenes with varying cover curvatures to study the extent of our proposed cover modeling approach. Detailed explanations of the dataset can be found in Appendix Sec. \ref{our_dataset}.

\textbf{Quantitative baseline comparisons.} 
We evaluate our method's novel-view synthesis capabilities on LLFF, FisheyeNeRF, EyefulTower, and CurvedCover datasets, ordered by ascending camera distortion levels. These datasets present increasing challenges, with the CurvedCover dataset posing the greatest difficulties due to its external cover and scene complexities. We assess visual quality using LPIPS \cite{lpips}, SSIM \cite{ssim}, and PSNR \cite{psnr}, comparing our method to NeRF, SCNeRF, NeuroLens, as shown in Table \ref{table_fisheye_eyeful} and \ref{table_curved_cover}.

\begin{table}[htbp]
    \centering
    \caption{\textbf{Quantitative comparison on the LLFF dataset.} See Sec \ref{evaluation_section} for an analysis of the performance.}

    \begin{tabular}{ccccc}
    \toprule  
    Scene & \centering{Model} & PSNR $\uparrow$& SSIM $\uparrow$& LPIPS $\downarrow$ \\ \hline
    \multirow{ 2}{*}{Flower} & NeRF& 32.2 & 0.937 & 0.067\\
    & Ours& 33.1 & 0.951 & 0.062 \\ \hline
    \multirow{ 2}{*}{Trex} & NeRF& 31.4 & 0.955 & 0.099\\
    & Ours& 32.3 & 0.963 & 0.094 \\
     \hline
    \end{tabular}
    \label{llff}
    \vspace{-4mm}
\end{table}

Our method consistently outperforms these baselines on scenes with greater distortions (FisheyeNeRF, EyefulTower, and CurvedCover). These comparisons highlight the effectiveness of our holistic camera distortion modeling compared to other camera-calibrating synthesis frameworks. To further assess and ensure that our method does not hinder novel-view synthesis quality with sufficiently calibrated datasets, we compare its results with the baselines on the LLFF dataset, as shown in Table \ref{llff}. Notably, our method does not hinder the innate novel view synthesis capability as the cover geometry essentially resorts back to a flat surface when the camera does not have a curved cover and is sufficiently calibrated. Finally, we support the aforementioned decomposed geometric Refractive Field learning objective as in Table \ref{quantitative_ablate}. For more details on the implementation and experiments on the 3DGS model, see Appendix Sec. \ref{3dgs_implementation} and Sec. \ref{3dgs_evaluation}.

\textbf{Qualitative evaluations on distorted captures.}
We compared our method with NeRF, SCNeRF and NeuroLens on both synthetic and real curved cover datasets as shown in Fig. \ref{main_qualitative}. The vanilla NeRF was unable to synthesize distorted captures, as consistent with previous findings in the literature \cite{Jeong2021Self-CalibratingFields}. Our approach achieves superior rendering quality due to its ability to explicitly model the curved cover geometry. This is particularly evident in scenes with significant refractive distortions, where the baseline methods struggle to accurately capture the light behavior, as shown in Fig. \ref{blowup}. With the modeled Refractive Field, the cover geometry can be recovered as visualized in Fig. \ref{surface}.

\textbf{Refractive Field representations.}
\begin{table}[htbp]
    \centering
    \caption{Ablation studies on the warmup process, Refractive Field, deformation field, and generalizability to the 3D-GS model.}
    \begin{tabular}{c@{\extracolsep{\fill}}ccc}
    \toprule  
    \centering{Model} & PSNR $\uparrow$& SSIM $\uparrow$& LPIPS $\downarrow$ \\ \hline
    ours-NeRF& 32.9 & 0.904 & 0.287\\
     ours-GS & 33.4 & 0.896 & 0.262 \\
     $\boldsymbol{-}$ warm up & 25.3 & 0.786 & 0.442 \\
    $\boldsymbol{-}$ Refractive Field & 29.4 & 0.861 & 0.307 \\
     \hline
    \end{tabular}
    \label{quantitative_ablate}
    \vspace{-4mm}
\end{table}
We conduct ablation studies on the synthetic ``Coffee Table'' sequence in the CurvedCover dataset to dissect the impact of the warm-up phase and Refractive Field, as detailed in Table \ref{quantitative_ablate}. End-to-end training solely on distortion parameters highlighted the need for a warm-up phase of $6000$ iterations, to avoid numerical instabilities. To probe the necessity for explicit geometric estimation of the Refractive Field, we directly output the ray offsets by modifying the Refractive Field's output layer to produce $\Delta \boldsymbol{r}$. Our findings confirm that the geometric Refractive Field is crucial for accurate light modeling. It alleviates the network's burden of simultaneously learning both the cover geometry and the physics of refraction, leading to more efficient and effective scene reconstruction.

\section{Limitations and Future Work}
Our method is primarily limited in its ability to handle specular effects from the protective cover since the Refraction Field only accounts for the primary refracted path for each ray. The specular reflections are present when ambient lights are reflected on the covers. Scenes with strong specular effects are generally not well-captured by NeRF frameworks without explicitly extending its ray tracing to capture higher-order light paths from the ambient environment, as discussed in Eikonal Field \cite{bemana2022eikonal}. For enabling specular reflections in splatting methods, additional attributes to handle specular surfaces could be amended to the 3D Gaussian color parameterization. 
\section{Conclusion}
\label{sec:conclusion}
In this study, we introduce SynthCover to enable novel view synthesis with sequences aberrated by protective covers, a common design choice adopted by extended reality headsets and field robots for durability. Our approach fits into existing novel view synthesis pipelines, enabling them to model sequences with cover-induced optical aberrations. Alongside this, we present a new dataset featuring both synthetic and real captures with aberrative covers to facilitate further research on this topic. Our evaluations, utilizing our dataset and existing novel view synthesis datasets, validate our model's broad applicability across various cover geometries. This underscores its potential as a distortion rendering tool for future robotic novel view synthesis applications. SynthCover is currently limited to handle constant material dispersion with monochromatic lights, without consideration of specular effects from the cover. For future work, we plan on extending our work to handle other optic effects caused by the cover such as specular reflections and scattering for more general applications. 
\newpage
{\small
\bibliographystyle{ieee_fullname}
\bibliography{references2}
}

\end{document}